\documentclass[10pt,twocolumn,letterpaper]{article}
\newcommand{\modelname}{PSDiffusion}
\usepackage{iccv}      
\usepackage{multirow}
\definecolor{iccvblue}{rgb}{0.21,0.49,0.74}
\usepackage[pagebackref,breaklinks,colorlinks,allcolors=iccvblue]{hyperref}

\def\paperID{13} 
\def\confName{ICCV}
\def\confYear{2025}

\title{PSDiffusion: Harmonized Multi-Layer Image Generation via\\ Layout and Appearance Alignment}

\author{
\large
Dingbang Huang$^{1}$ \quad
Wenbo Li$^{1}$ \quad
Yifei Zhao$^{1}$ \quad
Xinyu Pan$^{2}$\\[0.2em]
\large
Chun Wang$^{3}$ \quad
Yanhong Zeng$^{4}$ \quad
Bo Dai$^{5\ast}$\\[0.5em]
\small
$^{1}$Shanghai Jiao Tong University \quad
$^{2}$The Chinese University of Hong Kong\\[0.2em]
\small
$^{3}$Zhejiang University \quad
$^{4}$Shanghai AI Laboratory \quad
$^{5}$The University of Hong Kong\\[0.2em]
\footnotesize
}

\begin{document}
\maketitle
\begin{abstract}

Transparent image layer generation plays a significant role in digital art and design workflows. Existing methods typically decompose transparent layers from a single RGB image using a set of tools or generate multiple transparent layers sequentially. Despite some promising results, these methods often limit their ability to model global layout, physically plausible interactions, and visual effects such as shadows and reflections with high alpha quality due to limited shared global context among layers. 
To address this issue, we propose \textbf{PSDiffusion}, a unified diffusion framework that leverages image composition priors from pre-trained image diffusion model for simultaneous multi-layer text-to-image generation. Specifically, our method introduces a global layer interaction mechanism to generate layered images collaboratively, ensuring both individual layer quality and coherent spatial and visual relationships across layers. 
We include extensive experiments on benchmark datasets to demonstrate that PSDiffusion is able to outperform existing methods in generating multi-layer images with plausible structure and enhanced visual fidelity.

\end{abstract}
    
\begin{figure}
\centering
\includegraphics[width=0.49\textwidth]{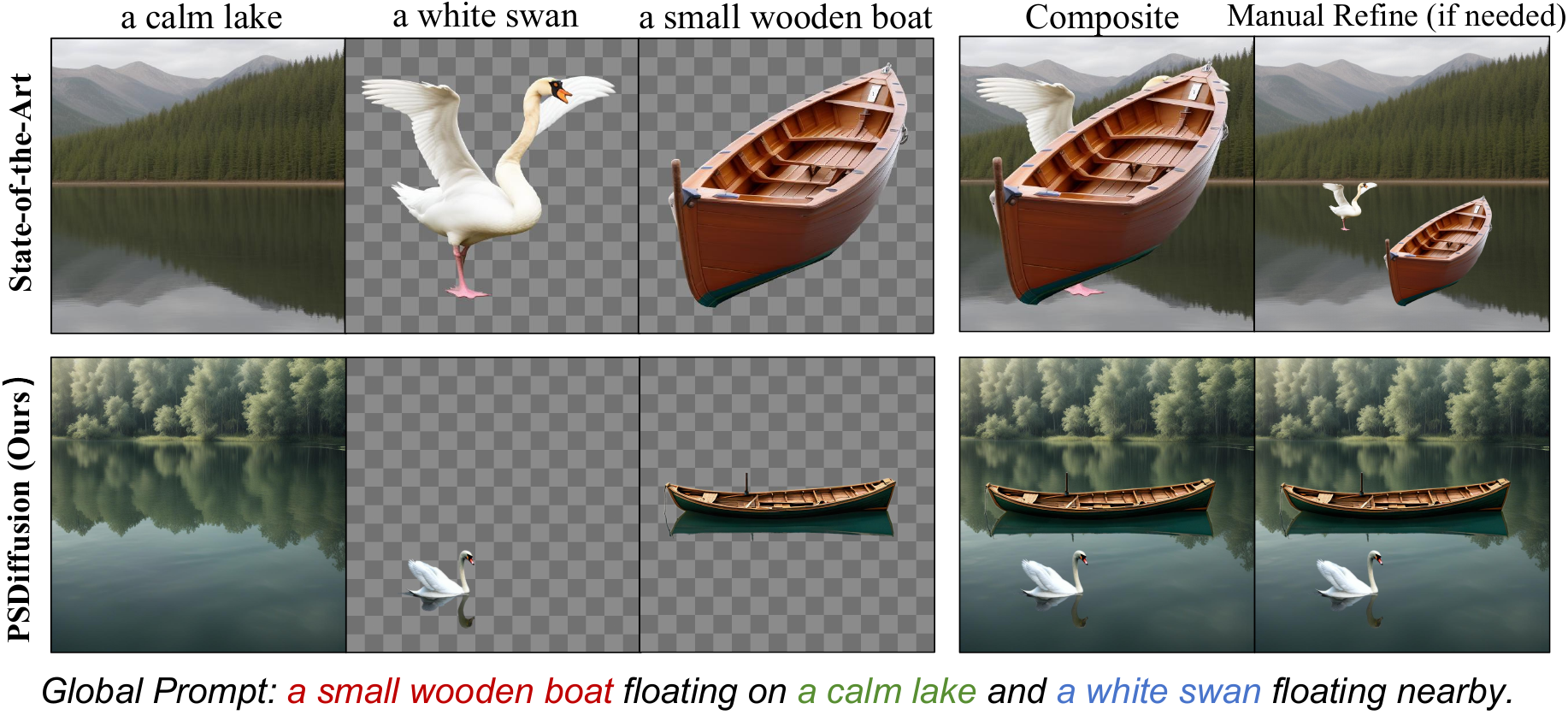}
\caption{
Given a complex prompt, our model, \modelname, produces plausible per-layer layouts for each instance and achieves harmonious inter-layer interactions. Compared to the state-of-the-art~\cite{zhang2024transparent}, \modelname\ generates layered images with more plausible spatial arrangement and more consistent appearances with realistic visual effects (e.g., reflections for the boat and swan), without manual spatial refinement. [Best viewed with zoom-in]}
\label{fig:teaser}
\end{figure}
\section{Introduction}
\label{sec:intro}

Diffusion models~\cite{rombach2022high,dalle3,imagen} have revolutionized image synthesis by generating high-fidelity, diverse visuals from textual prompts, but they are limited to producing images as unified, single-layer outputs which are not easy to edit. In contrast, images with layered representations provide human designers with great advantages in image manipulations, such as precise editing through element isolation, compositional flexibility via layer recombination, and collaborative iteration through asset sharing.

This limitation of pretrained generative models catalyzed the emergence of RGBA image layer synthesis. Text2Layer~\cite{Text2Layer} explored generating image layers through segmentation guidance. LayerDiffuse's~\cite{zhang2024transparent} ``latent transparency" mechanism accelerated the field, which enabled pretrained LDMs to produce high-fidelity single transparent layers. Building on transparent single-layer image generation, recent works have made strides toward multi-layer synthesis. LayerDiffuse~\cite{zhang2024transparent} itself extended its architecture for sequential background-to-foreground generation . Fontanella \etal~\cite{fontanellagenerating} explored separate multiple foreground layers generation and post-composition. LayerDiff~\cite{huang2024layerdiff} attempted to generate multi-layer images simultaneously, but it can only generate isolated, non-overlapping layers with binary masks. These paradigms mainly depend on straightforward layer stacking instead of considering multiple layers in a holistic manner. Therefore, they tend to neglect the interactions among layers, including comprehensive layout, physics-plausible contacts and visual effects like shadows and reflections, while these are essential to cohesive multi-layer generation and editing.

Another critical challenge underlying this mult-layer image generation stems from the scarcity of high-quality multi-layer RGBA image datasets.  While text-image pairs abound in datasets like LAION~\cite{schuhmann2021laion400mopendatasetclipfiltered}, commercially restricted access to premium transparent assets severely limits usable multi-layer RGBA data. Existing open-source multi-layer datasets exhibit complementary weaknesses. MAGICK~\cite{Burgert_2024_CVPR} and  Multi-layer Dataset~\cite{zhang2024transparent} provide high quality layers but are constrained to 1-2 layers per composition, whereas MuLAn~\cite{tudosiu2024mulan} supports 2-6 layers per image but suffers from poor alpha-matte quality because it relies on automated post-processing pipelines.

To overcome these dataset limitations, we propose the Inter-Layer Dataset, which consists of 30K meticulously curated multi-layer images, each containing 3-6 layers with artist-grade alpha mattes and rich layer interactions. Recognizing that automated synthesis and post-editing inherently fail to guarantee precise alpha channels and contextually coherent inpainting, we engaged about 35 professional designers to construct this dataset through a human-in-the-loop image composition workflow. Each composition operation involves spatial layout optimization, physics-aware layer interactions, and global stylistic consistency.

Although existing pretrained text-to-image generation models can only generate unified single layer images, given a global text prompt describing the background, multiple foregrounds and their interactions, they did demonstrate prominent capability to spontaneously arrange the layout of multiple elements, ensure natural interactions between constituent subjects, and maintain global harmony and consistency in generated images. We believe that these pretrained text-to-image generation models can provide interaction priors for multi-layer generation. Building on this fact, we propose PSDiffusion, a unified diffusion framework that utilizes the interaction priors from pretrained diffusion models, to simultaneously generate multi-layer images with coherent global structure and realistic layer interactions.  We design an attention-based global-layer interactive mechanism to model the layer interactions paradigm. For one thing, to achieve a harmonious global layout for foreground layers, we develop a layer cross-attention reweighting module, where we extract the cross-attention map from the global layer to guide the position of foreground layers. For another, to facilitate content coherence and interaction among different layers, and avoid layer entanglement, we develop a partial joint self-attention module, where we encourage content sharing from global layer across the foreground and background layers. 

Given a global prompt and various layer-specific prompts as input, our PSDiffusion model enables simultaneous multi-layer RGBA image generation with enhanced interactions among all layers. The layer-specific prompts can either be automatically decomposed by LLMs or manually defined by users for precise layer control. Furthermore, with the layered image representations, our model supports various layer-wise editing for users to implement precise layer control.
In summary, our contributions are as follows: 
 \begin{itemize}
\item We propose Inter-Layer, a high quality multi-layer RGBA image dataset consisting of 30K samples, each containing 3-6 layers with artist-grade alpha mattes and rich, harmonious layer interactions.
\item We present PSDiffusion, an end-to-end diffusion framework for simultaneous multi-layer image generation. Leveraging a global-layer interactive mechanism, our model generates layered images concurrently and collaboratively, ensuring not only high quality and completeness for each layer, but also spatial and visual interactions among layers for global coherence. 
\end{itemize}

\begin{figure*}[t]
    \centering
    \includegraphics[width=1\linewidth]{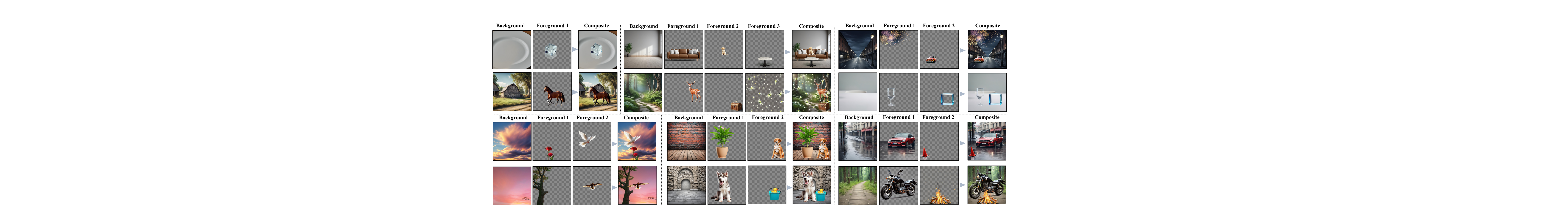}
    \caption{Visual results of \modelname. Through the design of Layer Cross-Attention Reweighting Module and Partial Joint Self-Attention Mechanism, \modelname\ synthesize multi-layer transparent images with plausible layout arrangements and harmonious appearance.}
    \label{fig:more-visual-results}
\end{figure*}
\section{Related Work}
\label{sec:related}

\subsection{Text-to-Image Generation}
Diffusion models have made significant advancements in image generation, particularly in text-guided image synthesis, leading to remarkable improvements in both the quality of generated images and the ability to follow complex prompts. The development of large-scale diffusion models trained on extensive datasets of paired text and images has established new benchmarks in this domain, with notable examples including Stable Diffusion~\cite{rombach2022high}, SDXL~\cite{podell2023sdxl}, Imagen~\cite{imagen}, and DALL-E 3~\cite{dalle3}. Furthermore, recent advancements in network architecture, especially with Diffusion Transformer networks (DiT) like Stable Diffusion 3~\cite{esser2024scaling} and Flux~\cite{flux2024} , have enhanced image fidelity and diversity. Moreover, other research efforts focus on integrating various control conditions into pre-trained image foundation models to enhance their controllability. For instance, Gligen~\cite{li2023gligen}, ControlNet~\cite{zhang2023control} and IP-Adapter~\cite{ye2023ip} offer greater flexibility in controllable image generation, allowing users to precisely manipulate aspects such as layout, depth, human pose, and style. However, these models can only generate a single RGB image during a single forward pass, rather than producing layered images, which are widely used formats that enable users to edit specific parts of an image without affecting other areas.

\subsection{Layered Image Generation}

\noindent\textbf{Image Layer Extraction with Post-RGB Process.}
Multi-layer image generation can be achieved primarily through two different paths. The first path, termed Post-RGB Process, operates through a sequential pipeline: (1) initial RGB image synthesis via diffusion models~\cite{rombach2022high}; (2) foreground extraction using open-vocabulary object detection and segmentation tools (\eg GroundingDINO~\cite{liu2024grounding},  SAM~\cite{Kirillov2023Segment}, Matting-Anything~\cite{li2023matting}); (3) background inpainting and foreground completion for occluded areas to derive multiple RGBA layers~\cite{rombach2022high, podell2023sdxl, yu2023inpaint, ao2024open}.  
Along this way, Tudosiuet \etal~\cite{tudosiu2024mulan} developed a training-free pipeline, which decomposes a monocular RGB image into a stack of RGBA layers comprising background and isolated instances, and released the MuLAn dataset. LayeringDiff~\cite{kang2025layeringdiff} begins by generating using an off-the-shelf image generative model, followed by disassembling the image into its constituent foreground and background layers. Yang \etal proposed LayerDecomp~\cite{yang_Layer_Decomposition}, a generative training framework that leverages both simulated and real-world data to enable layer-wise post-decomposition with visual effect representation.
Despite the operational validity of these methods, the multi-stage post-processing approach imposes a nontrivial computational redundancy and risks compounding errors through iterative operations—a limitation that frequently manifests itself as incoherent visual semantics or poor alpha quality. 

\noindent\textbf{Direct Transparent Image Layer Generation.}
RGBA direct generation focuses on synthesizing image layers directly, bypassing the traditional pipeline of generating a full image first and then extracting transparent layers. The earliest work, Text2Layer~\cite{Text2Layer}, extends the Stable Diffusion architecture into a dual-layer generation framework. It jointly synthesizes a background scene and semantically aligned foreground elements through coordinated latent space optimization.
Subsequently, LayerDiffuse~\cite{zhang2024transparent} introduces "latent transparency", enabling large-scale pretrained latent diffusion models to generate either single transparent images. This approach has gained a reputation for its high-quality single-layer transparency generation.
In parallel, Alfie~\cite{quattrini2024alfie} and Diffumatting~\cite{hu2024diffumatting} also develop different pipelines to generate single-layer RGBA images.

Building on single-layer generation capabilities, multi-layer synthesis is achieved through either sequential or separate generation paradigms. Specifically, LayerDiffuse~\cite{zhang2024transparent} adopts a sequential approach, proposing a background-conditioned transparent layer generation that iteratively synthesizes layers from background to foreground. Fontanella \etal~\cite{fontanellagenerating} and LayerFusion~\cite{dalva2024layerfusion} employ a separate paradigm, where individual foreground layers are generated independently and then composited.

Recently unified text-guided multi-layer generation has attracted increasing attention. Specifically, LayerDiff~\cite{huang2024layerdiff} proposes a simultaneous multi-layer synthesis framework using a unified diffusion model with 3D convolutions. However, it produces low-quality outputs limited to binary masks instead of full alpha channels, failing to handle overlapping foregrounds and visual effects which are essential for layer editing. ART~\cite{pu2025art} introduces an Anonymous Region Layout Planner, aligning visual tokens with text tokens to directly synthesize spatially coherent multi-layer images. However, it relies on precise bounding boxes provided by users or generated by LLMs and its anonymous layout mechanism results in layer entanglement issue, where semantically distinct elements (\eg, foreground characters and background props) are incorrectly merged into one layer, and cause issues on alpha quality.

\begin{figure}[t]
\centering
    \includegraphics[width=0.5\textwidth]{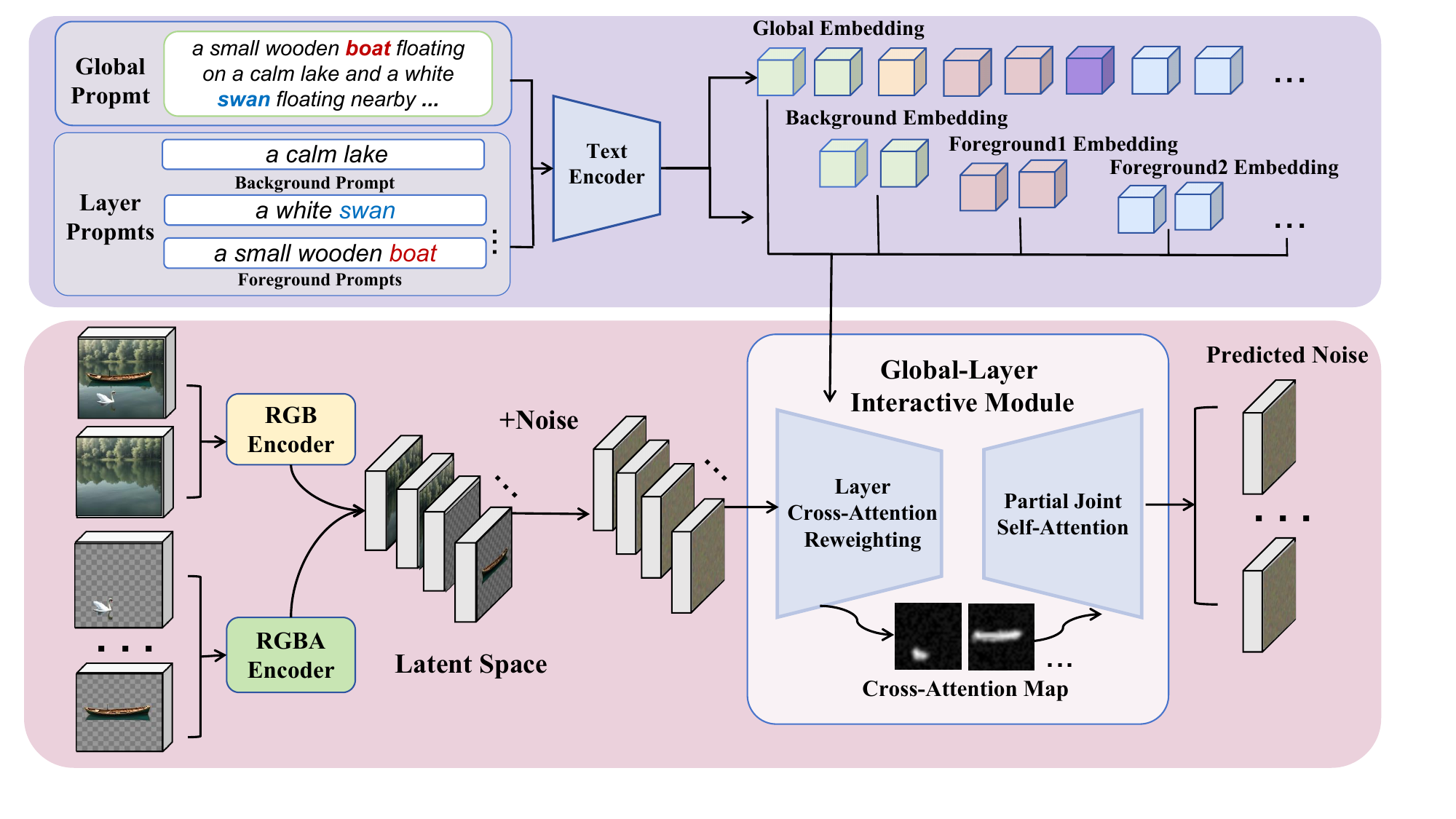}
    \caption{Architecture of PSDiffusion. Our framework processes multi-layer compositions through three key components: (1) A transparent VAE encoder preserving alpha channels; (2) Layer cross-attention reweighting for layout plausibility; (3) Partial joint self-attention for inter-layer context modeling.}
    \label{fig:pipeline}
\end{figure}
\section{Methodology}
\subsection{Problem Formulation}

A multi-layer composable image is defined as a structure containing a background layer $B = I_c^B \in \mathbb{R}^{H \times W \times 3}$ in RGB format and $K$ foreground layers $\{F_k\}_{k=1}^K$ in RGBA format, where each $F_k = \left(I_c^{F_k}, I_\alpha^{F_k}\right)$ consists of color channels $I_c^{F_k} \in \mathbb{R}^{H \times W \times 3}$ and an alpha channel $I_\alpha^{F_k} \in \mathbb{R}^{H \times W \times 1}$. Following the premultiplied alpha convention, each foreground is preprocessed as $F'_k = I_c^{F_k} \ast I_\alpha^{F_k}$ through element-wise multiplication. 

The composition pipeline initializes with the background $\smash{I_0 = I_c^B}$, then recursively blends layers via alpha operations. The $k$-th composite image is given by
\begin{align}
I_k = \left(1 - I_\alpha^{F_{k}}\right) \cdot I_{k - 1} + I_\alpha^{F_{k}} \ast I_c^{F_{k}},\ k \in [K]
\end{align}
where $F_{k}$ denotes the $k$-th foreground. The final composite image $I_{K}$ is gained after $K$ iterations. This mathematical formulation explicitly captures two fundamental mechanisms: (1) The multiplicative attenuation of background visibility through successive transparency operations;  (2) The nested modulation of foreground layer contributions based on depth-ordered occlusions.

\subsection{PSDiffusion}

Existing multi-layer image generation models contend with inherent challenges in synthesizing coherent compositions and fostering harmonious interactions between distinct high-quality transparent foreground layers and the background image~\cite{tudosiu2024mulan,zhang2024transparent}. However, it has been observed that pretrained image foundation models demonstrate proficiency in generating intricate and visually coherent compositions, effectively integrating various objects and backgrounds within a single RGB image, owing to their training on extensive datasets derived from real-world imagery.
Our primary insight is to exploit the layout and interaction priors from the denoising process of the global RGB image, to guide the generation of foreground and background layers.

To fully take advantage of the layout arrangement and appearance harmonization of the global RGB models, we propose a novel framework, denoted as \modelname, which concurrently generates multiple layers in a single feedforward pass, ensuring both reasonable layouts and harmonized interactions among the layers. As delineated in Figure~\ref{fig:pipeline}, \modelname\ comprises three principal components. First, \modelname\ finetunes a pretrained RGB image Variational Autoencoder (VAE) into an RGBA VAE, thereby facilitating the generation of the alpha channel requisite for RGBA images, as articulated by LayerDiffuse~\cite{zhang2024transparent}. Second, we introduce a layer cross-attention reweighting mechanism that extracts spatial layout information from the global image containing comprehensive content, generated via a global prompting strategy. This design paradigm ensures a coherent arrangement of disparate foreground objects. Finally, to promote harmonious interactions among the various foreground layers, we implement a partial joint self-attention module for inter-layer context modeling. 

\begin{figure*}[t]
\centering
    \includegraphics[width=1\textwidth]{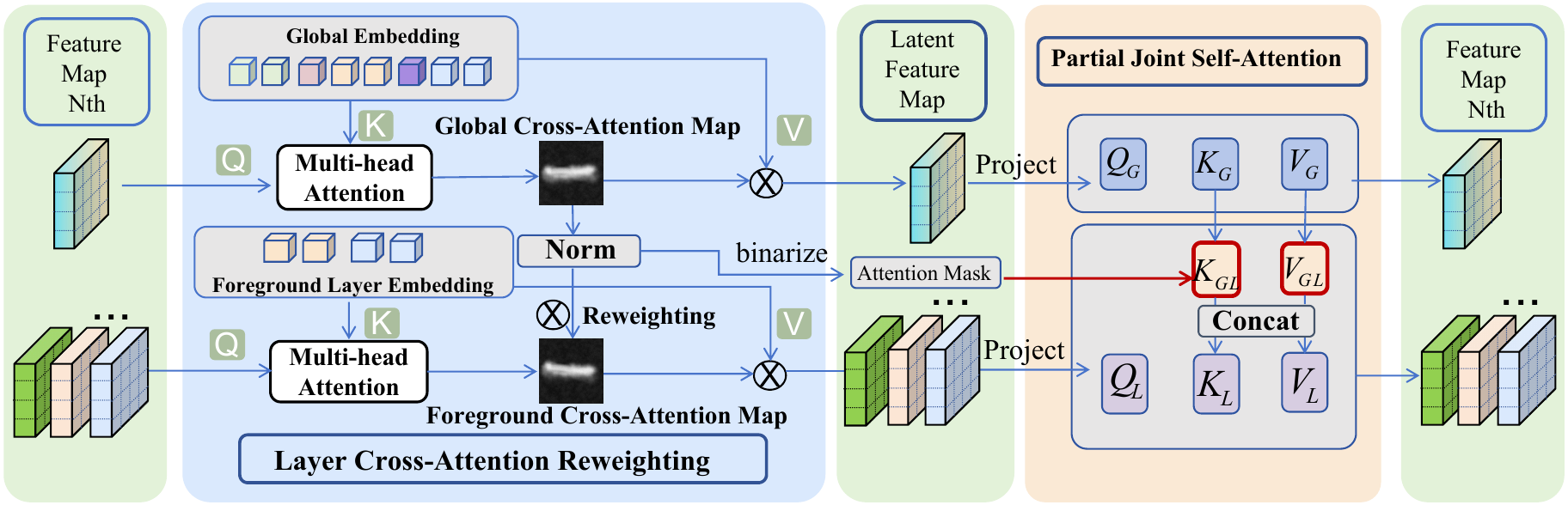}
    \caption{Overview of our global-layer interactive mechanism, composed of layer cross-attention reweighting module and partial joint self-attention module. Layer cross-attention reweighting module extracts the cross-attention map from the global branch to reweight the cross-attention map of the layer branch, guiding the position of foreground layers. Partial joint self-attention module implements a shared attention across global branch and layer branch to facilitate context-aware feature modeling. }
    \label{fig:global}
\end{figure*}

\noindent\textbf{Layer Cross-Attention Reweighting.}
To ensure a reasonable layout arrangement of foreground layers, we propose a novel approach that diverges from recursive layer generation. Our key insight is to refer to the layout arrangement from the denoising process of a complete RGB image generated using the global prompt. As illustrated in Figure~\ref{fig:global}, we utilize the cross-attention block of pre-trained image foundation model to extract the layout priors. Leveraging the emergent correspondence between prompts and features—a phenomenon widely demonstrated and utilized in existing literature~\cite{hertzprompt,tang2023emergent,cao_2023_masactrl,epstein2023diffusion,hu2023anomalydiffusion}—we can precisely derive and control the location of  foreground instance by calculating and manipulating the cross-attention map associated with the foreground instance token. As shown in Figure~\ref{fig:pipeline}, we select the token of the colored noun representing the foreground subject because its attention map most prominently displays the location of the foreground subject. 

At timestep $t$ in the $n$-th cross-attention block, the global text embedding $\psi(p)$ is projected to a key matrix $\mathbf{K}^n = \ell_\mathbf{K}^n(\psi(p))$ and a value matrix $\mathbf{V}^n = \ell_\mathbf{V}^n(\psi(p))$ , and the deep spatial features of the noisy image $\phi(z_t)$ is projected to a query matrix $\mathbf{Q}^n=\ell_\mathbf{Q}^n(\phi(z^t))$, via learned linear projections $\ell_\mathbf{K}^n$, $\ell_\mathbf{V}^n$, and $\ell_\mathbf{Q}^n$, respectively. The global attention map $\mathbf{M}^{\mathrm{g}}\in \mathbb{R}^{hw\times s} $can be calculated by
\begin{align}
    \mathbf{M}^{\mathrm{g}} = \mathrm{Softmax}\left(\dfrac{\mathbf{Q}^n\left(\mathbf{K}^n\right)^\mathrm{T}}{\sqrt{d}}\right).
\end{align}
We extend our method to obtain the layer attention map $M^{l}\in \mathbb{R}^{hw\times s}$ in the latent feature of the transparent image layer using an analogous calculation approach, where $hw$ denotes the size of the latent feature and $s$ denotes the length of the tokens. We extract the specific attention-map of the token representing the foreground subject from the global map and layer map, denoted as $\mathbf{M}_i^{\mathrm{g}} \in \mathbb{R}^{hw}$ and $\mathbf{M}_i^{\mathrm{l}} \in \mathbb{R}^{hw}$. Since there are differences between the global map $\mathbf{M}_i^{\mathrm{g}}$ and layer map $\mathbf{M}_i^{\mathrm{l}}$ during the original denoising process, we utilize the global map as a guide to reweight the layer map. We implement this layer cross attention map reweighting by
\begin{align}
\tilde{\mathbf{M}}_i^{\mathrm{l}} = \mathrm{Norm}\left(\mathbf{M}_i^{\mathrm{g}}\right) \odot \mathbf{M}_i^{\mathrm{l}},\\ \hat{\mathbf{M}}_i^{\mathrm{l}} = \tilde{\mathbf{M}}_i^{\mathrm{l}} \cdot \dfrac{\max\left(\mathbf{M}_i^{\mathrm{l}}\right)}{\max\left(\tilde{\mathbf{M}}_i^{\mathrm{l}}\right)}.
\end{align}
where $\mathrm{Norm}(\cdot)$ denotes Min-Max normalization on the whole feature dimension,  $\odot$ denotes Hadamard Product, $\max(\cdot)$ is applied on the feature dimension as well for stability,. 
This process provides a more precise attention map that clearly delineates the position and shape of areas associated with the foreground layer prompt. The global attention map is then strategically processed and integrated into the self-attention blocks of the latent diffusion model. Through this approach, we achieve a more nuanced and controlled layout arrangement for multi-layer image generation.

\noindent\textbf{Partial Joint Self-attention.}
While our proposed layer cross-attention reweighting mechanism
ensures a reasonable layout arrangement for each layer,
challenges remain in harmonizing the color, style and physics interactions across different layers. To address the appearance inconsistency among layers, we introduce a novel partial joint self-attention mechanism for cross-layer feature context modeling. This approach facilitates more nuanced interactions among layers, ultimately achieving consistent color, geometry and lighting conditions within the composite image.
As illustrated in Figure~\ref{fig:global}, we concatenate the features of the global branch into the $n$-th self-attention block of the layer branch, 
\begin{align}
\mathbf{Q} = \mathbf{Q}_L = \mathbf{W}^n_\mathbf{Q} z_i
\end{align}
\begin{align}
 \mathbf{K} = \mathbf{W}^n_\mathbf{K} \left[\mathbf{K}_{GL}, \mathbf{K}_L\right],\ \mathbf{V} = \mathbf{W}^n_\mathbf{V} \left[\mathbf{V}_{GL}, \mathbf{V}_L\right]
\end{align}

where $z_i$ denotes the latent feature for the $i$-th transparent layer, $[\cdot,\cdot]$ denotes the concatenation operation. $\mathbf{K}_{GL}$ and $\mathbf{V}_{GL}$ are calculated by 
\begin{align}
\mathbf{K}_{GL} = \{ \mathbf{K}_G \cdot \mathrm{Mask}_i^\mathrm{g} \},\ \mathbf{V}_{GL} = \{\mathbf{V}_G \cdot \mathrm{Mask}_i^\mathrm{g}\}.
\end{align}
where $\mathrm{Mask}_i^\mathrm{g}$ is derived from the global attention map $\mathbf{M}_i^\mathrm{g}$ in the Layer Cross-Attention Reweighting block and applied with filtering, blurring, and binarization, $\{\cdot\}$ denotes that we apply the attention mask in the multi-head attention. With these masks, we can restrict the $i$-th layer to query context information only from the corresponding and nearby regions in the global image, thus avoiding layer entanglement issues. Meanwhile, due to the inherent appearance harmony and element interactions of the global image, the content sharing from the associate regions of global branch to layer branch is enough for the composite harmonization. 

Overall, such a design enables the model to refer to the appearance features of the global layer, promoting a more cohesive and harmonized image generation across different foreground transparent layers and background layer.

\begin{figure}
    \centering
    \includegraphics[width=1\linewidth]{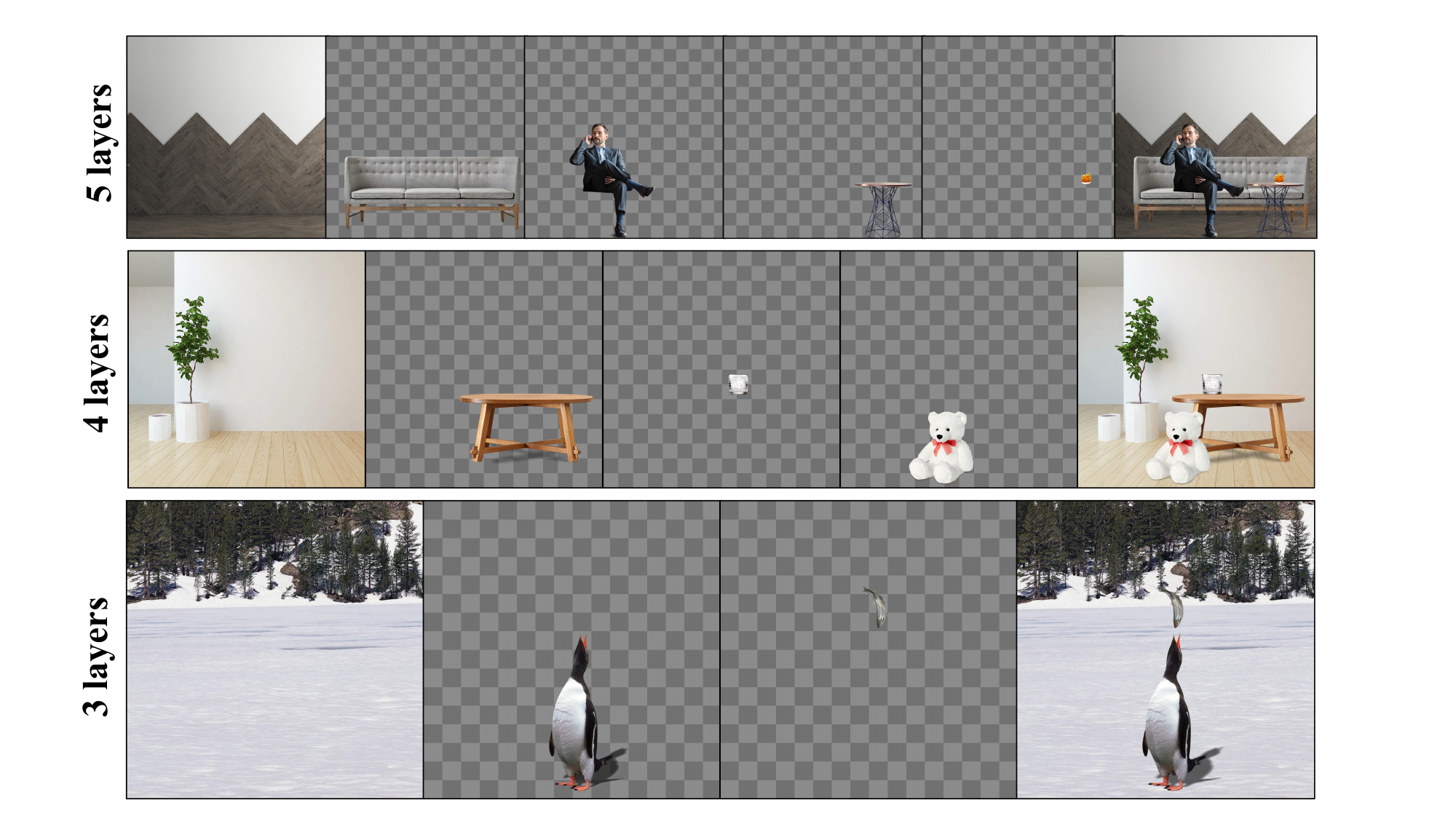}
    \caption{Examples of our proposed Inter-Layer dataset.
    }
    \label{dv_main}
\end{figure}

\subsection{Inter-Layer Dataset}
To address the limitations of existing multi-layer datasets, we developed the dataset \textbf{Inter-Layer} through a human-centric curation pipeline. This dataset comprises 30K high-fidelity multi-layer compositions, each containing 3 - 6 layers with professional-grade alpha mattes and sophisticated inter-layer relationships. Recognizing the inadequacy of automated synthesis for achieving precise transparency channels and contextual coherence, we engaged professional designers in a multi-stage workflow. Curators systematically curated assets from diverse sources including web repositories, BG20K~\cite{li2022bridging}, MAGICK~\cite{Burgert_2024_CVPR}, LayerDiffuse-generated transparent assets~\cite{zhang2024transparent} and Flux-generated assets~\cite{flux2024}, followed by rigorous manual refinement. Designers optimized each composition through: (1) Spatial layout organization with an aesthetic consideration; (2) Physics-plausible interactions among all layers, especially geometry and orientation consistency. Designers strategically select layer combinations from our extensive RGBA asset library whose spatial and geometry relationships are pre-aligned. Minor 2D rotational and shape adjustments are then applied to optimize spatial coherence and guarantee physics-plausible contacts. (3) Global visual harmonization of all elements. Designers perform comprehensive visual adjustments to maintain consistency between foreground objects and their contextual environment. This includes calibrating saturation, hue, contrast, resolution, sharpness and brightness to match surrounding conditions. Additionally, shadows and reflections are manually crafted according to environmental factors (e.g., water surfaces, lighting direction) to achieve photorealistic integration.  This human-in-the-loop approach ensures both technical precision in alpha channel quality and visual consistency in multi-layer storytelling, significantly advancing beyond existing datasets' capabilities in layer complexity and visual authenticity. 
Figure~\ref{dv_main} presents examples from our proposed dataset, highlighting its high quality, diverse multi-foreground layers, harmonized compositions, and consistent appearance.

\begin{figure*}[t]
    \centering
    \includegraphics[width=0.95\textwidth]{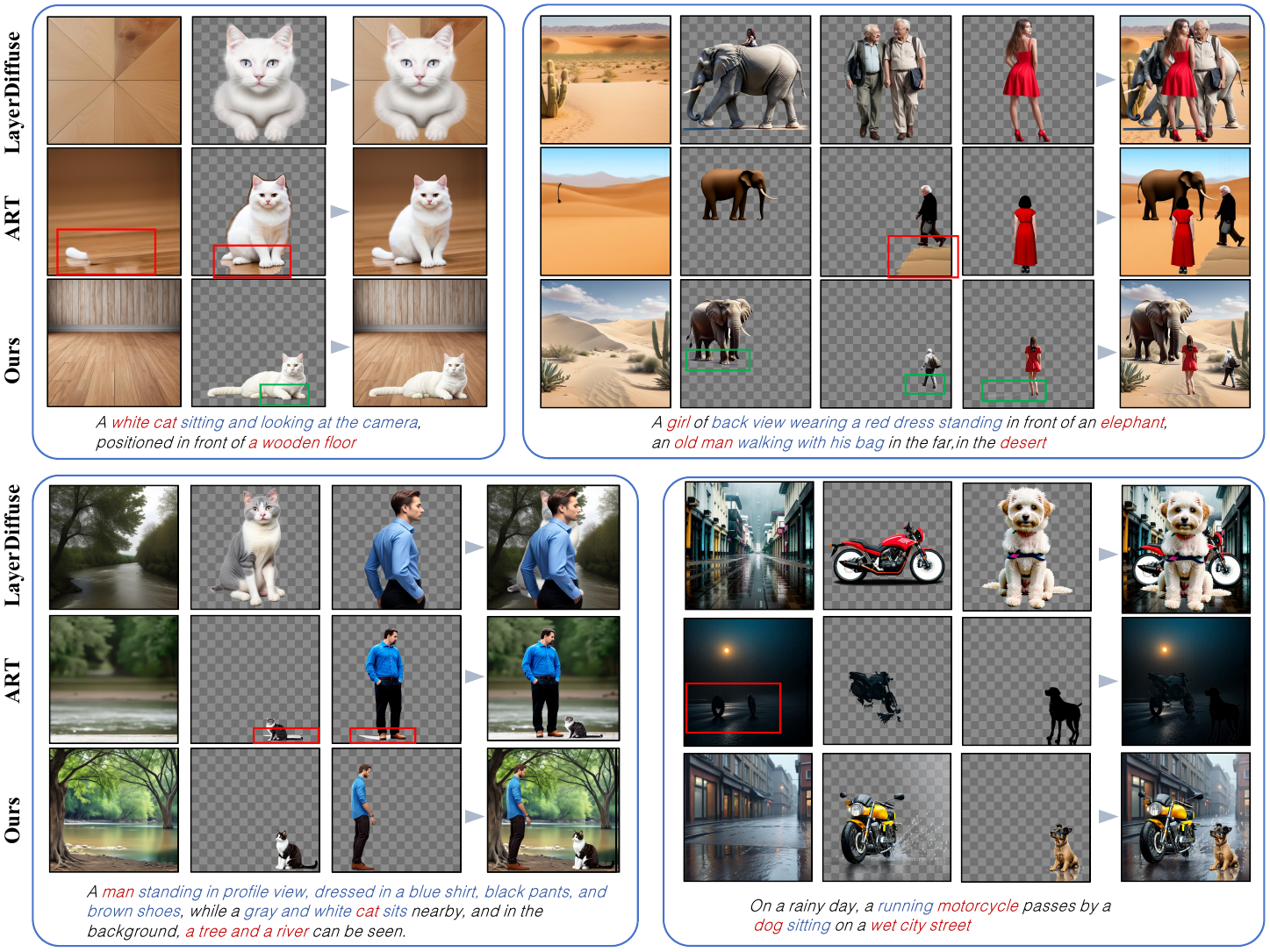}
    \caption{Qualitative comparison with state-of-the-art methods. Compared to existing methods, our model synthesizes multi-layer images with reasonable layout arrangement, harmonious appearance with visual effects and high alpha quality. Zoom in to check the details of foreground and composite images like shadows, reflections and foreground edges. As exemplified by the multi-layer ensemble in the upper-right quadrant, the shadows of the elephant, man and woman are consistent with the background in their respective positions.}
    \label{fig:Qualitative comparison}
\end{figure*}

\section{Experiments}
\subsection{Implementation Details}
To better preserve the interaction priors of existing models, PSDiffusion is fine-tuned from Stable Diffusion XL~\cite{podell2023sdxl} using LoRA~\cite{hu2022lora}. For global layer, we trained the additional $\mathbf{K}_{GL}$,$\mathbf{V}_{GL}$  linear layers in all self-attention blocks for joint attention. For foreground layers and background layers, we trained $\mathbf{Q}$,$\mathbf{K}$,$\mathbf{V}$,$\mathbf{Out}$ linear layers in all attention blocks. Layer Cross-Attention Reweighting is implemented in layers at the resolution of 16, and the Partial Joint Self-attention is applied to all self-attention layers. LoRA rank is set to 256. We use the pre-trained VAE from Stable Diffusion XL for the global and background branch, and fine-tune it into an RGBA VAE to support the alpha channel of RGBA foreground image following LayerDiffuse~\cite{zhang2024transparent}. During training, all layers of a multi-layer sample are encoded into latent space and added with the same noise at the same timestep. We use the AdamW optimizer and set the learning rate at $1e\text{-}4$. Training is carried out over our Inter-Layer dataset with a total batch size of 8 on 60 NVIDIA A800 GPU hours for 10,000 iterations.

\subsection{Compared with State-of-the-Art}
\label{subsec:sota}

\begin{table*}[t]
  \centering
  \small
  \begin{tabular}{@{} c  cccc  c  cc @{}}
    \toprule
    \multirow{2}{*}{\textbf{Method}}
      & \multicolumn{4}{c}{Composite Image}
      & & \multicolumn{2}{c}{Layer Image} \\
    \cmidrule(lr){2-5} \cmidrule(lr){7-8}
      & CLIP Score$\uparrow$ & FID$\downarrow$ 
      & Layout Harm.$\uparrow$ & Inter.Plaus.$\uparrow$
      & & CLIP Score$\uparrow$ & FID$\downarrow$ \\
    \midrule
    LayerDiffuse~\cite{zhang2024transparent}
      & 29.77 & 89.12 & 0.265 & 0.214 
      & & 31.25 & 85.85 \\
    ART~\cite{pu2025art}
      & 29.93 & 96.54 & 0.743 & 0.702 
      & & --    &128.1  \\
    \modelname (ours)
      & \textbf{31.89}& \textbf{87.32}
      & \textbf{0.766}&\textbf{0.751}
      & & \textbf{31.76}&\textbf{83.71} \\
    \bottomrule
  \end{tabular}
  \caption{Quantitative comparison with state-of-the-art methods. We report performance on both composite images and layer images in terms of visual quality, text alignment and composite layout harmony and interaction plausibility.}
  \label{tab:quantitative-result}
\end{table*}

 We compare our model with LayerDiffuse~\cite{zhang2024transparent} and the latest work ART~\cite{pu2025art}, as they are the state-of-the-art layered image synthesis methods, and the only methods that release the source code. For evaluation, we construct 5,000 sets of multi-layer prompts, each composed of three foreground prompts, one background prompt and a global prompt describing their spatial and semantic interactions. We use vision-language model~\cite{Qwen2.5-VL} to generate the prompts to establish diverse evaluation scenarios.\\
We conduct quantitative evaluation using multiple metrics. For image quality assessment, we measure feature distribution discrepancies between synthesized layers and reference datasets via FID~\cite{heusel2017gans} (Fréchet Inception Distance). Specifically, the RGBA layer is evaluated against the MAGICK dataset~\cite{Burgert_2024_CVPR}, while the composite images are compared with the COCO dataset~\cite{lin2014microsoft}. To assess text-image alignment for each layer, we employ the CLIP Score metric~\cite{hessel2021clipscore}, which quantifies the similarity between image features and text prompts. To evaluate the harmony of the global layout and the plausibility of the layer interactions, we utilize vision-language model~\cite{Qwen2.5-VL} to conduct scoring comprehensively. The definition of layout harmony includes the rationality of the global layout and the scale of each element. The definition of interaction plausibility includes physical contacts and geometry consistency among entities, coordination of color and style, and consistency of light and shadows.

\noindent\textbf{Qualitative Results.}
In Figure~\ref{fig:Qualitative comparison}, we present the qualitative results on text prompt multi-layer image generation.  LayerDiffuse~\cite{zhang2024transparent} tends to stack all layers at the center of the image, neglecting the spatial interactions among all layers. The latest work ART showcases reasonable layout but relies on the precise bounding boxes given by users. It also exhibits the problem of imperfect alpha quality and layer entanglement. In contrast, our model displays not only high quality of RGBA layers, but also the capability to arrange the layers to reasonable scale and position, and maintaining the visual interactions among the layers. Zoom in to see the details of foreground and composite images like edges, shadows and reflections. More qualitative results of our model are shown in Figure~\ref{fig:more-visual-results}.

\noindent\textbf{Quantitative Results.}
We report quantitative results in Table~\ref{tab:quantitative-result}. We compare our \modelname with the state-of-the-art multi-layer image generation models LayerDiffuse~\cite{zhang2024transparent} and ART~\cite{pu2025art} on composite images. For LayerDiffuse, we implement the proposed sequential background-to-foreground generation to gain the multi-layer images. For ART, since it relies on the bounding-boxes as input, we first apply the LLM layout generation proposed by the authors given our global prompt, and then use the generated bounding boxes and global prompts to generate the multi-layer images.  As shown in the table, our PSDiffusion outperforms LayerDiffuse and ART across four metrics. The results on layout harmony and interaction plausibility directly showcase our strengths on generating multi-layer image with enhanced layer interaction and global rationality. Furthermore, for the alpha image layer, the results in Table~\ref{tab:quantitative-result} shows that our PSDiffusion also achieves excellent alpha image quality and better text alignment.

\noindent\textbf{User Study.}
A comprehensive perceptual evaluation was conducted with 30 participants recruited from academic institutions to holistically evaluate the visual synthesis performance of multi-layer image generation models. The study encompassed totally 171 visual samples derived from 15 distinct multi-layer text prompts, with each case containing three components: foreground layers, backgrounds layers and composite outputs generated by two baseline methods alongside our proposed \modelname. Each sample underwent multi-criteria evaluation using 5-point Likert scales (1: deficient, 5: exceptional) focusing on three critical dimensions: (1) semantic-textual congruence measuring prompt relevance and contextual appropriateness, (2) visual-perceptual fidelity evaluating aesthetic coherence and artifact suppression, and (3) structural clarity assessing image definition and detail preservation. The comparative evaluation results are summarized in Table~\ref{tab:commands}, demonstrating that our method consistently achieves higher user ratings across than other methods. 

\subsection{Ablation Studies}
\begin{figure}
    \centering
    \includegraphics[width=0.99\linewidth]{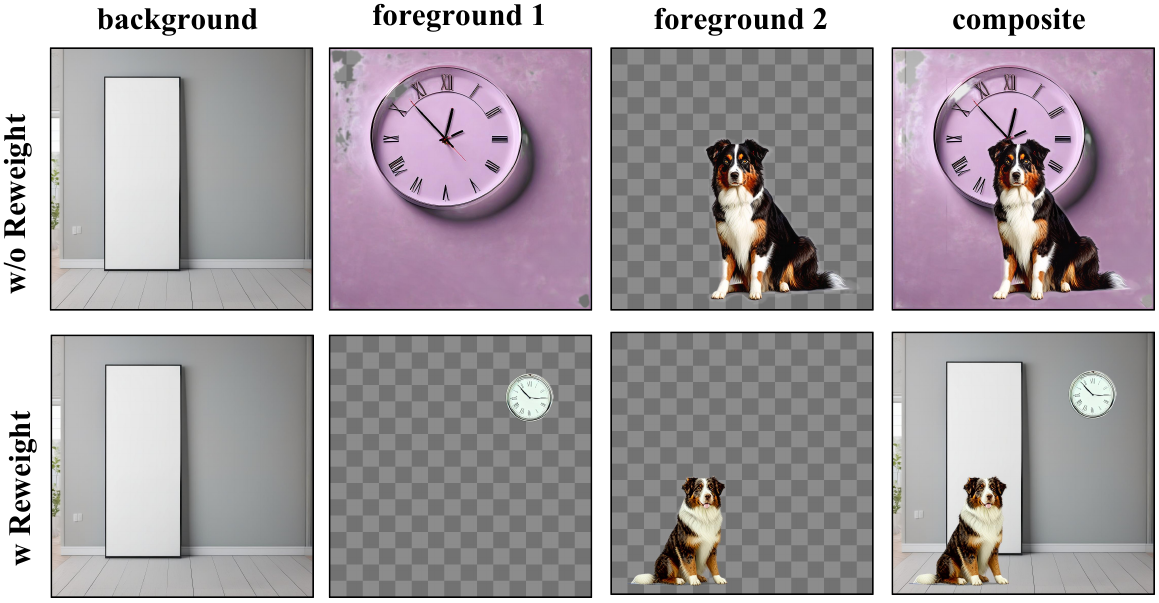}
    \caption{Visual results of ablation study for layer cross-attention reweighting. Full PSDiffusion exhibits more plausible spatial layout including postion and scale.}
    \label{reweighting}
\end{figure}

\noindent\textbf{Layer Cross-Attention Reweighting.}
In this module, we utilize the cross-attention map of the global layer, and implement the reweighting operation on the cross attention map of the foreground layers to control the global layout.  As shown in Figure~\ref{reweighting}, without the layer attention-map reweighting, the foreground object tends to occupy the most of the image like LayerDiffuse and even shows poor quality on the alpha channel, especially for the foreground that should have been small in the composite image. 

\begin{figure}
    \centering
    \includegraphics[width=0.98\linewidth]{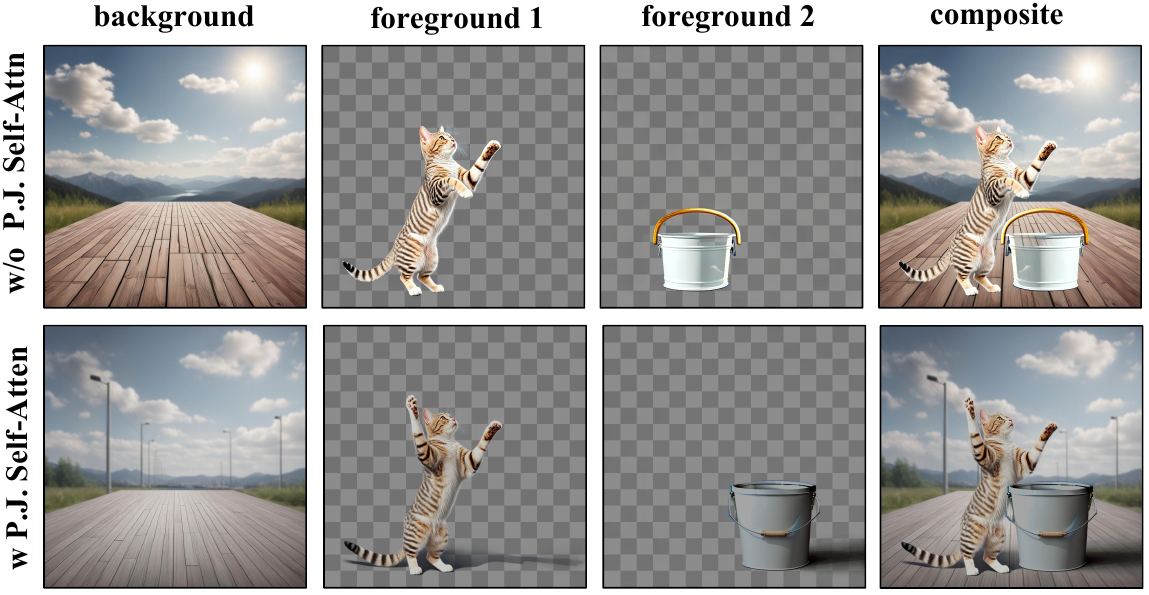}
    \caption{Visual results of ablation study for partial joint self-attention. Full \modelname\ demonstrates more coherent appearances and visual effects (\eg, shadows).}
    \vspace{6px}
    \label{fig:abaltion w/o l}
\end{figure}

\noindent\textbf{Partial Joint Self-Attention.}
In our partial joint self-attention blocks, we implement the content sharing from global layer across foreground and background layers to encourage layer interaction plausibility like the shadows.  Figure~\ref{fig:abaltion w/o l} displays that removing this block leads to visual effects like  rigid stacks of separated foreground layers hence global disharmony.

As shown in Table~\ref{tab:ablation}, removing either partial joint self-attention block or layer cross-attention reweighting block results in significant decline in both layout harmony and layer interactions. This showcases their entangled importance to the coherence and interaction plausibility of the composite image.

\begin{table}[t]
  \centering
  \small
  \setlength{\tabcolsep}{4pt}
  \begin{tabular}{cccc}
    \toprule
    Method & Foreground$\uparrow$ & Background$\uparrow$ & Composite$\uparrow$ \\
    \midrule
    LayerDiffuse~\cite{zhang2024transparent} & 1.64 & 3.63 & 1.53 \\
    ART~\cite{pu2025art}                    & 1.30 & 2.53 & 3.14 \\
    \modelname(ours)                       & \textbf{2.06} & \textbf{3.83} & \textbf{3.35} \\
    \bottomrule
  \end{tabular}
  \caption{User study on multi-layer image generation models. Our \modelname\ are the most popular with users.}
  \label{tab:commands}
\end{table}

\begin{table}[t]
\centering
\small
\setlength{\tabcolsep}{2pt}
  \begin{tabular}{ccccc}
    \toprule
    Model & CLIP$\uparrow$ & FID$\downarrow$ & Layout Harm.$\uparrow$ & Inter.Plaus.$\uparrow$ \\
    \midrule
    w/o Reweighting & 31.64 & 89.18 & 0.565 & 0.418 \\
    w/o Joint Atten. & 30.71 & 89.54 & 0.488 & 0.350 \\
    Full model & \textbf{31.89} & \textbf{87.32} & \textbf{0.766} & \textbf{0.751} \\
    \bottomrule
  \end{tabular}
  \caption{Ablation study on Layer Cross-Attention Reweighting and Partial Joint Self-Attention on composite images}
  \label{tab:ablation}
\end{table}

\section{Conclusion}
In this paper, we propose the Inter-Layer Dataset, a high-quality multi-layer RGBA image dataset with artist-grade alpha mattes and rich layer interactions. Building on this, we introduce PSDiffusion, an end-to-end diffusion framework for simultaneous multi-layer image generation. We design a global-layer interactive mechanism to extract interaction priors from pre-trained diffusion models.
Our model generates layered images concurrently and collaboratively, ensuring not only high quality and completeness for each layer, but also spatial and visual interactions among layers for global coherence.

\textit{Future Work. }
Given that our method leverages the intrinsic interaction priors of pre-trained generative models and achieves excellent multi-layer generation results through an attention-based global-layer interactive  module, it holds the potential to be migrated to other attention-based pre-trained generative models. In the future, we plan to explore extending our method to a broader range of generative models, including DiT-based image generation~\cite{esser2024scaling,flux2024}, video generation~\cite{wan2025,blattmann2023stable}, and 3D generation models~\cite{lin2023magic3d}.
{\small \bibliographystyle{ieeenat_fullname} \bibliography{main}
}

\end{document}